\documentclass{article}
\usepackage{ijcai16}

\usepackage{times}
\usepackage{bm}
\usepackage{algorithmic}
\usepackage{algorithm}
\usepackage{amsfonts}
\usepackage{xspace}
\usepackage{amsmath}
\usepackage{amsthm}
\usepackage{graphicx}
\usepackage{epsfig}
\usepackage{subfigure}
\usepackage{graphicx}
\graphicspath{{figs/}}

\theoremstyle{break}
\newtheorem{Cor}{Corollary}
\theoremstyle{plain}

\usepackage{color}
\usepackage{hyperref}

\theoremstyle{marginbreak}

\theoremstyle{change}

\newtheorem{Prop}[Cor]{Proposition}

\def\x{{\bf x}}
\def\y{{\bf y}}
\def\w{{\bf w}}

\def\bb{\bm{\beta}}

\def\fs{f}
\def\ys{{{\mathrm{y}}}}

\def\R{{\mathbb{R}}}

\def\one{{\bf 1}}

\def\alphas{{\mathrm{\alpha}}}
\def\alphab{{\bm{\alpha}}}

\def\betas{{\bf \beta}}

\def\lambdas{{\bf \lambda}}

\def\K{{\bf K}}

\def\tK{\tilde{{\bf K}}}

\def\vb{{\bf v}}

\def\I{{\bf I}}

\def\Q{{\bf Q}}

\def\x{{\bf x}}
\def\w{{\bf w}}

\def\bb{\bm{\beta}}

\def\R{{{R}}}

\def\DomainX{{\mathcal{X}}}
\def\DomainY{{\mathcal{Y}}}
\def\DomainZ{{\mathcal{Z}}}

\def\x{{\bf x}}
\def\y{{\bf y}}
\def\w{{\bf w}}

\def\bb{\bm{\beta}}

\def\fs{f}

\def\R{{\mathbb{R}}}

\def\one{{\bf 1}}

\def\alphas{{\bf \alpha}}
\def\alphab{{\bm \alpha}}

\def\betas{{\bf \beta}}

\def\zb{{\bf{z}}}

\def\K{{\bf K}}

\def\tK{\tilde{{\bf K}}}

\def\I{{\bf I}}

\def\Q{{\bf Q}}

\def\x{{\bf x}}
\def\y{{\bf y}}
\def\w{{\bf w}}

\def\bb{\bm{\beta}}

\def\z{{\bf z}}

\def\1{{\mathbf 1}}
\def\0{{\mathbf 0}}

\def\Rr{{R}}

\def\gs{{g}}

\title{Simple and Efficient Learning using Privileged Information}

\author{Xinxing Xu$^{1}$, Joey Tianyi Zhou$^{1}$, Ivor W. Tsang$^{2}$, Zheng Qin$^{1}$, Rick Siow Mong Goh$^{1}$ \and Yong Liu$^{1}$\\
$^{1}$ Institute of High Performance Computing, A*STAR, Singapore  \\
$^{2}$ University of Technology Sydney\\
\{xuxinx, zhouty, qinz, gohsm, liuyong\}@ihpc.a-star.edu.sg, ivor.tsang@uts.edu.au}

\begin{document}

\maketitle

\begin{abstract}
  The Support Vector Machine using Privileged Information (SVM+) has been proposed to train a classifier to utilize the additional privileged information that is only available in the training phase but not available in the test phase.
  In this work, we propose an efficient solution for SVM+ by simply utilizing the squared hinge loss instead of the hinge loss as in the existing SVM+ formulation, which interestingly leads to a dual form with less variables and in the same form with the dual of the standard SVM.
  The proposed algorithm is utilized to leverage the additional web knowledge that is only available during training for the image categorization tasks. The extensive experimental results on both Caltech101 and WebQueries datasets show that our proposed method can achieve a factor of up to hundred times speedup with the comparable accuracy when compared with the existing SVM+ method.
\end{abstract}

\section{Introduction}
In traditional machine learning paradigm, the classifiers are trained based on the features of training data, and the test is done using the same type of features.
However, in real-world applications, there will be additional information that is only available during training but not available during test.
For example, for the image categorization problem, we are usually classifying the images from different categories.
Take the Caltech101 data set as an example, there are a total number of 101 object categories in the training set,
and each of the object category has its associated descriptions, such as the textual descriptions from Wikipedia.
The descriptions about each concept can be associated to each training image simply by using its label information, and hence they are only available at the training
stage, but not available during the test phase. Another example is the learning from weakly labeled web images. The images in the Web are usually associated with descriptions from tags that are uploaded by the users. The tag information can be collected during training data collection.
However, during test we may only have the test images that do not contain any descriptions.

The aforementioned two examples fall into the new learning paradigm of the Learning using Privileged Information (LUPI)~\cite{SVM+},
in which the additional information is referred to as the privileged information.
The Support Vector Machine using Privileged Information (SVM+)~\cite{SVM+} has been proposed for utilizing the privileged information.
As the hinge loss is used in~\cite{SVM+}, we refer to the formulation in their work as SVM1+ in the following.
It has been proved theoretically that the incorporating of the additional privileged information can improve the
convergence rate~\cite{PechyonyV10}. The SVM+ attracts much attention recently~\cite{vapnik15b,PIDist,lapin2014learning}, and has been successfully applied to different applications~\cite{MDAHSPlus,RankTransfer,FouadTRS12}. Following this learning scenario, some recent works proposed to utilize the privileged information for the different learning scenarios such as learning to rank \cite{RankTransfer}, Gaussian Process classifier~\cite{GPCPI}, clustering~\cite{FeyereislA12} and distance metric learning \cite{metricLearningPlus,ITML+}.

However, the proposed optimization method for SVM1+ in~\cite{SVM+} is based on its dual form. If there are a total number of $n$ training data, the
corresponding dual problem is a Quadratic Programming (QP) problem with $2n$ variables, which is 2 times larger than the dual form of the standard SVM~\cite{SVM} with only $n$ variables.
Besides, as the constraints in the dual form of SVM+ are different with those in the dual form of the standard SVM, the efficient implementations and the off-the-shelf solvers for SVM such as LibSVM~\cite{libsvm} and SVMLight~\cite{svmlight} can not be readily utilized, and the additional efforts must be spent to design the specific solvers for optimization of SVM1+~\cite{SMOSVM+}.

To overcome the aforementioned problem, in this work, we study the optimization of SVM+. Specifically, we propose a new objective function called Support Vector Machine using Privileged Information with Squared Hinge Loss (SVM2+) by simply replacing the hinge loss in SVM1+ with the squared hinge loss. The dual form of SVM2+ is a Quadratic Programming (QP) problem with only $n$ variables, which significantly reduces the training complexity.
What's more, the dual form of SVM2+ shares the same form with the dual of the standard SVM. Therefore, the existing off-the-shelf QP solvers for SVM can be readily applied to solve the proposed objective function.

We apply our proposed approach to the image categorization tasks by leveraging the additional information from web knowledge.
We evaluate our algorithms on Caltech101 dataset with the additional textual descriptions for the training data obtained from Wikipedia, as well as the WebQueries dataset with the additional tag information. The extensive empirical results show that our proposed SVM2+ can achieve 110.6 (\emph{resp.} 92.5) times speedup on Caltech101 (\emph{resp.}WebQueries) when compared with the solution for SVM1+ as in~\cite{SVM+} yet with comparable classification accuracies. Hence, the experimental evaluation demonstrates both the efficiency and the effectiveness of our proposed SVM2+ algorithm for utilizing the additional web knowledge as privileged information for the task of image categorization.

In the following, we firstly introduce the SVM1+ in Section~\ref{sec:LUPI}.
In Section~\ref{sec:SVM2+}, we propose the objective function and the solution for SVM2+.
The experimental results are shown in Section~\ref{sec:experiment} and finally the conclusions are given in Section~\ref{sec:conclusion}.

\section{Support Vector Machine using Privileged Information with Hinge Loss}\label{sec:LUPI}
For the classical machine learning paradigm, we are given a set of $n$ independent and identically distributed (IID) training pairs
$\{(\x_1,\ys_1),\ldots,(\x_n,\ys_n)\}$ with $\x_i \in \DomainX \subset \R^{d}$ and $\ys_i \in \DomainY \subset \R$ that are generated according to a fixed but unknown probability distribution   $P(\x,\ys)$. The target of any machine learning algorithm is therefore to learn a function $\fs(\x)$ that can make the risk functional $\Rr(\fs) = \frac{1}{2}\int | \ys - f(\x) |\mathrm{d}P(\x,\ys)$ minimized. Different from the classical learning paradigm, the Learning using Privileged Information (LUPI) paradigm tackles the learning senario where an additional ``teacher" is only available in the training phase but not available during the test phase. Specifically, we are given a set of $n$ independent and identically distributed training triplets $\{(\x_1,\z_1,\ys_1),\ldots,(\x_n,\z_n,\ys_n)\}$ with $\x_i \in \DomainX \subset \R^{d}$, $\z_i \in \DomainZ \subset \R^{s}$ and $\ys_i \in \DomainY \subset \R$ that are generated according to a fixed but unknown probability distribution $P(\x,\z, \ys)$. In the test phase, the aim is the same with that of the classical learning paradigm, \emph{i.e.}, classifying any test samples $\x_{i}$'s, where $ i =n+1,\ldots,n+m$. The $\DomainX$ is referred to as the decision space as the test is done only in $\DomainX$, while $\DomainZ$ is referred to as the correcting space~\cite{PechyonyV10}.

The support vector machine using privileged information (SVM+) has been proposed in~\cite{SVM+} for utilizing the privileged information.
Specifically, $\fs(\x) = \w' \x + b$ is the decision function based on the main feature $\x$ and $\gs(\zb) = \zb' \vb + \rho$ is the correcting function based on the privileged information $\zb$, and the SVM+ with hinge loss (SVM1+) is proposed to optimize the following objective function~\cite{SVM+}:
\begin{eqnarray}\label{eqn:SVM+L1}
\min_{\w,\vb, b,\rho,\xi_i}  \!\!\!&& \!\!\! \frac{1}{2}||\w||^{2} + C \sum_{i=1}^{n}\xi_i + \frac{\lambdas}{2} ||\vb||^{2}   \nonumber \\
\mbox{s.t.},  \!\!\!&& \!\!\! \zb_i' \vb +  \rho = \xi_i, \nonumber \\
\!\!\!&& \!\!\! \ys_i \left( \w' \x_i + b \right) \geq 1 - \xi_i, \\
\!\!\!&& \!\!\! \xi_i \geq 0.\nonumber
\end{eqnarray}
The objective function in (\ref{eqn:SVM+L1}) is solved in its dual form as in~\cite{SVM+,vapnik15b}. Specifically, by introducing the non-negative multipliers $\alphab = [\alphas_1,\ldots,\alphas_n]'$ and $\bb = [\betas_1,\ldots,\beta_n]'$, the dual form is given as in the following Quadratic Programming (QP) problem:
\begin{eqnarray}\label{eqn:SVM+L1_dual}
\max_{\alphab,\bb} \!\!\!&&\!\!\! \alphab'\one - \frac{1}{2} (\alphab\odot\y)' \K(\alphab\odot\y) \nonumber \\
 \!\!\!&&\!\!\! - \frac{1}{2\lambdas}(\alphab + \bb - C\one)' \tK (\alphab + \bb - C\one)\nonumber \\
\mbox{s.t.}, \!\!\!&&\!\!\!  \sum_{i=1}^{n} (\alphas_i + \betas_i -C ) = 0 , \\
\!\!\!&&\!\!\! \sum_{i=1}^{n} \alphas_i \ys_i = 0, \nonumber\\
\!\!\!&&\!\!\! \alphas_i \geq 0, \betas_i \geq 0, \nonumber
\end{eqnarray}
where $\y = [\ys_1,\ldots,\ys_n]'\in \R^{n}$ is the label vector, $\one = [1,\ldots,1]'\in R^{n}$, $\odot$ is the elementwise product between two vectors/matrices, $\K \in \R^{n\times n}$ with $\K_{ij} = \x_i'\x_j$ is the kernel matrix obtained from the main feature $\x$, while $\tK \in \R^{n\times n}$ with $\tK_{ij} = \zb_i'\zb_j$ is the kernel matrix constructed using the privileged information $\zb$. Although it is linear kernel here, any type of non-linear kernel~\cite{SMMKL} can be readily utilized in (\ref{eqn:SVM+L1_dual}).

Note that the dual problem in (\ref{eqn:SVM+L1_dual}) is a standard $2n\times 2n$ QP problem, and it can be solved by the general QP solver.
However, the $2n \times 2n$ problem in (\ref{eqn:SVM+L1_dual}) takes more memory and requires more training time than the original SVM dual form in a $n\times n$ QP problem. Moreover, due to the introduced constraints for the $\alphab$ and $\bb$, the existing efficient solvers for the SVM (\emph{e.g.}, LibSVM and SVMLight) cannot
be readily applied to solve (\ref{eqn:SVM+L1_dual}). Therefore, additional efforts are required for the development of the efficient solution and the softwares such as~\cite{SMOSVM+}.

In the following, we propose the SVM+ with squared hinge loss, and we are interested to see that we can obtain a QP problem in size of $n\times n$ that is also in the same form with the standard dual form of SVM and can therefore be solved using any off-the-shelf efficient QP solver developed for SVM.

\section{Support Vector Machine using Privileged Information with Squared Hinge Loss}\label{sec:SVM2+}
In order to solve the SVM+ more efficiently and motivated by the using of squared hinge loss for SVM~\cite{IntroSVM}, we propose the Support Vector Machine using Privileged Information with Squared Hinge Loss (SVM2+). We learn the decision function $\fs(\x) = \w' \x + b$ as well as the correcting function $\gs(\zb) = \vb' \zb_i$\footnote{We augment an additional 1 to the feature vector $\zb$ to incorporate the bias term, and thus $\zb$ should be replaced with $[\zb;1]\in\R^{s+1}$, but we still use $\zb$ to represent the augmented feature vector for ease of presentation.} by optimizing the following objective function:
\begin{eqnarray}\label{eqn:SVM+L2}
\min_{\w,\vb,b,\xi_i} \!\!\!&&\!\!\! \frac{1}{2}||\w||^{2} + \frac{C}{2} \sum_{i=1}^{n}\xi_i^{2} + \frac{\lambdas}{2} ||\vb||^{2}   \nonumber \\
\mbox{s.t.}, \!\!\!&&\!\!\! \vb' \zb_i  = \xi_i,  \\
\!\!\!&&\!\!\! \ys_i \left( \w' \x_i + b \right) \geq 1 - \xi_i, \nonumber
\end{eqnarray}
where $C$ and $\lambdas$ are still the two regularization parameters.
In the objective function (\ref{eqn:SVM+L2}), we just simply replace the original hinge loss (\emph{i.e.}, $ \sum_{i=1}^{n}\xi_i$) in (\ref{eqn:SVM+L1}) with the squared hinge loss (\emph{i.e.}, $ \sum_{i=1}^{n}\xi_i^{2}$).
The improvement looks quite simple and straightforward, but we can observe the significant differences and the benefits in the dual form as shown in the following
proposition \ref{prop:SVM+2_dual}.
\begin{Prop}\label{prop:SVM+2_dual}
The dual form of the optimization problem in (\ref{eqn:SVM+L2}) is given as in the following form:
\begin{eqnarray}\label{eqn:SVM+_2_qp}
\min_{\alphab} && \frac{1}{2} (\alphab\odot\y)'\left(\K + \Q_{\lambdas}\odot (\y\y')  \right)(\alphab \odot\y) - \alphab' \one, \nonumber \\
\mbox{s.t.} && \alphab \geq 0,  \alphab'\y = 0,
\end{eqnarray}
where $\y = [\ys_1,\ldots,\ys_n]'\in \R^{n}$ is the label vector, $\one = [1,\ldots,1]'\in R^{n}$, $\odot$ is the elementwise product between any two vectors/matrices, $\alphab = [\alphas_1,\ldots,\alphas_n]'\in \R^{n}$ is the vector of the Lagrangian multipliers, $\K \in \R^{n\times n}$ with $\K_{ij} = \x_i'\x_j$ is the kernel matrix constructed from $\x$, and $\Q_{\lambdas}\in \R^{n\times n}$ is a deformed kernel matrix in the form of
\begin{eqnarray}\label{eqn:Qker}
\Q_{\lambdas} = \frac{1}{\lambdas} \left( \tK -  \tK \left(\frac{\lambdas}{C}\I_{n} + \tK \right)^{-1} \tK \right),
\end{eqnarray}
where $\tK \in \R^{n\times n}$ with $\tK_{ij} = \z_i'\z_j$ is the kernel matrix constructed from $\z$.
\end{Prop}
Note that similarly for SVM1+, we just use the linear kernel as a demonstration, but any type of non-linear kernel can be readily utilized in (\ref{eqn:SVM+_2_qp}).

Interestingly, we observe that the optimization problem in (\ref{eqn:SVM+_2_qp}) is just a $n\times n$ Quadratic Programming (QP) problem.
We only need to optimize with respect to the $n$ dual variables $\alphab$ rather than the $2n$ dual variables as in (\ref{eqn:SVM+L1_dual}).
The reduced dual problem can not only save memory for optimization but also reduce the computational complexity.
Although there is an additional matrix inversion operation as in (\ref{eqn:Qker}), it can be done very efficiently when compared with
the optimization of the QP problem. We reduce the size of the dual form by utilizing the squared hinge loss instead of the hinge loss.

More importantly, the QP problem in (\ref{eqn:SVM+_2_qp}) also shares the same form with the QP problem of that of the classical SVM~\cite{SVM}. The difference is the changing of the kernel matrix as in the QP problem. Specifically, we can just replace the kernel matrix $\K$ in the original SVM to be $\K + \Q_{\lambdas}\odot (\y\y')$.
More interestingly, the matrix $\Q_{\lambdas}$ is simply a transformation of the kernel matrix $\tK$ constructed on the privileged information by using (\ref{eqn:Qker}).
The QP problem in (\ref{eqn:SVM+_2_qp}) can be readily solved by any existing solvers (\emph{e.g.}, LibSVM) specifically developed for the standard SVM.

\section{Experiments}\label{sec:experiment}
In this section, we show the experimental results of our proposed algorithm and the baseline algorithms for image categorization tasks on the Caltech101~\cite{caltech101} and the WebQueries~\cite{webquery} datasets.

We compare our proposed new algorithm SVM2+ with the baseline algorithms, \emph{i.e.}, SVM, SVM-2K~\cite{SVM2K} and SVM1+.
The SVM is trained based only on the visual feature extracted from images, and the SVM2K is a two-view learning algorithm that trains two classifiers on the two views simultaneously, and
we only use the view from visual feature for prediction. Therefore, the SVM2K, SVM1+ and SVM2+ all utilized the additional privileged information during training.
However, only the images are used for the test for all the algorithms as the privileged information is not available in our learning setting.
For all the algorithms, we set all the regularization parameters in the range of $\{10^{-3},10^{-2},\ldots,10^{3}\}$.
The best result from each algorithm is reported for performance comparisons.
\subsection{Dataset Description and Feature Extraction}
Table~\ref{tab:data_sets} summarizes the details of the used two datasets for our experiments, which are described in the following.
\subsubsection{Caltech101}
The Caltech101 dataset\footnote{http://www.vision.caltech.edu/Image\_Datasets/Caltech101} contains images from 101 object categories (\emph{e.g.}, ``helicopter'', ``elephant'' and ``chair'' etc.) and a background category that contains the images not from the 101 object categories.
For each object category, there are about 40 to 800 images, while most classes have about 50 images.
The resolution of the image is roughly about 300$\times$200 pixels.
Following the popular settings~\cite{caltech101}, we utilize 10 images per class for training and up to 50 images per category for the test.
Finally, we get 1020 images in the training set, and 2995 images in the test set.

\subsubsection{WebQueries}
The WebQueries dataset\footnote{https://lear.inrialpes.fr/$\sim$krapac/webqueries/webqueries.html} is composed of 71,478 images obtained by retrieving a total number of 353 textual web queries (\emph{e.g.},``eiffel tower", ``violin" and ``France flag" etc.).
For each image in the WebQueries dataset, there are corresponding textual descriptions either in English or French.
Besides, the relevant labels have been annotated by human manually. In our experiments, the images with English queries and
the textual queries with more than 100 images according to the ground truth labels are used as the evaluation queries.
In this way, we obtain a total number of 76 queries for the final classification tasks.
For each of the 76 queries, we used 10 (\emph{resp.}, 50) images for constructing the training set (\emph{resp.}, test set).
Therefore, we have 760 images for training, while we have 3800 images for test.

\subsubsection{Image Feature Extraction}
For the image representation, the deep learning features are extracted from each of the images due to its excellent performance for computer vision tasks~\cite{DeCAF}.
The MatConvNet~\cite{MatConvNet} is used to extract the deep learning features. The vgg-s model~\cite{Chatfield14} is used in our work. It is pre-trained on the 1.2 million ImageNet dataset~\cite{ImageNetData}, and the 4096-dimensional output of the fc6 in the deep Convolutional Neural Network (CNN) model from each image is employed as the visual feature representation. The same type of visual features are extracted for both the Caltech101 and the WebQueries datasets.

\subsubsection{Web Knowledge as Privileged Information}
For the Caltech101, it is difficult to obtain descriptions for each of the image.
We therefore discover the web knowledge by searching the category name of each category from Wikipedia.
Then we collect the text descriptions of each concept from the webpage of Wikipedia.
Obtaining the textual descriptions for all the categories, we further use the term frequency-inverse document frequency (TF-IDF) to convert each textual description into the bag of word frequency feature vector. We finally get a 29,535 dimensional feature vector for each of the object category. During training, each training image is associated with one vector that
represents its object category as its privileged information. As we do not have the ground-truth labels for test images, the privileged information is
not available during the test phase.

\begin{table}[t]
\centering
\setlength{\tabcolsep}{3pt}
\caption{A summarization of the datasets used in our experiments. The n and m are the number of training images and test images, and \#c is the number of total classes, and d and s are the feature dimensions for the main feature and privileged feature, respectively.}\label{tab:data_sets}
\begin{tabular}{|c|c|c|c|c|c|c|c|c|c|c|c|c|c|c|}
\hline
 & n & m & \#c  & d & s  \\
\hline
\hline
Caltech101 & 1020 & 2995 & 102 & 4096  & 29535\\
\hline
WebQueries & 760 &  3800 &  76 & 4096 & 2000\\
\hline
\end{tabular}
\end{table}

For the WebQueries dataset, each image is associated with a tag that contains a short description of the image in the Website.
We also collect the textual descriptions from each of the training images, and then we remove the stop-words and calculate the term-frequency (TF) for
each of the textual descriptions. The top-2000 words from the whole corpora are used as the vocabulary, and finally each textual description is represented as
a 2000-dimensional feature vector.
\begin{table}[t]
\centering
\setlength{\tabcolsep}{3pt}
\caption{The classification accuracies (\%) of all methods on the Caltech101 and the WebQueries datasets.}\label{tab:acc_all}
\begin{tabular}{|c|c|c|c|c|c|c|c|c|c|c|c|c|c|c|}
\hline
 & Caltech101 & WebQueries   \\
\hline
\hline
SVM & 85.41 & 53.74 \\
\hline
SVM2K & 84.82 &  53.21 \\
\hline
SVM1+ & 85.79 &  54.30  \\
\hline
SVM2+ & 86.20 &  54.43 \\
\hline
\end{tabular}
\end{table}

\subsection{Experimental Results and Discussion}

We use the linear kernel for both the visual feature and the privileged textual feature to train one-vs-others classifiers for each object category or query, and then we assign the labels of the test image to be the one with the highest output from classifiers on all categories or queries. The classification accuracy is used as the performance measurement.

The experimental results with the classification accuracies for the different algorithms are shown as in Table~\ref{tab:acc_all}.
We can observe that the additional privileged information does help the classification tasks for both
the datasets by using the LUPI framework.
Therefore, it is beneficial to utilize the web knowledge for the task of image categorization.
Besides, the SVM2K is worse than SVM and the other algorithms, which shows that
the training of the two-view classifiers is not effective for the LUPI learning paradigm.
We observe that the results from SVM2+ on the two datasets are slightly better than the results from SVM1+, which demonstrates
that it is more suitable to utilize the squared hinge loss for the SVM+ on these two applications.
Besides, the training speed differs significantly as shown later.

\begin{table}[t]
\centering
\setlength{\tabcolsep}{3pt}
\caption{The training CPU time (Seconds) of all methods on the Caltech101 and WebQueries datasets.}\label{tab:speed_all}
\begin{tabular}{|c|c|c|c|c|c|c|c|c|c|c|c|c|c|c|}
\hline
 & Caltech101  & WebQueries   \\
\hline
\hline
SVM & 17.86 & 1.997 \\
\hline
SVM2K & 732.8 &  80.79 \\
\hline
SVM1+ & 867.6 &  189.0  \\
\hline
SVM2+ & 7.847 &  2.044 \\
\hline
\end{tabular}
\end{table}

We compare the training CPU time of the different algorithms with a Lenovo desktop (3.20GHz CPU
with 8GB RAM) and Matlab implementation.
The parameters with the best classification accuracy for each algorithm are used to report the training time.
As the SVM1+ cannot be fit into the standard QP solver for SVM, we utilize the commercial software Mosek\footnote{https://www.mosek.com/} with academic license to solve the Quadratic Programming problem.
For SVM and SVM2+, we all utilize the LibSVM\footnote{https://www.csie.ntu.edu.tw/$\sim$cjlin/libsvm/} solver. \
The results in Table~\ref{tab:speed_all} show the CPU time of each algorithm on the whole Caltech101 and WebQueries datasets.
Table~\ref{tab:speed_all_times} shows the speedup times of our proposed SVM2+ when compared with the SVM1+.
We can observe from the table that our proposed SVM2+ achieves  110.6 and 92.5 times speedup respectively on Caltech101 and WebQueries datasets when compared with SVM1+.

\begin{table}[t]
\centering
\setlength{\tabcolsep}{3pt}
\caption{The relative speedup times of our proposed SVM2+ with respect to SVM1+ on the Caltech101 and WebQueries datasets.}\label{tab:speed_all_times}
\begin{tabular}{|c|c|c|c|c|c|c|c|c|c|c|c|c|c|c|}
\hline
 & Caltech101  & WebQueries   \\
\hline
\hline
SVM2+ & 110.6 &  92.5 \\
\hline
\end{tabular}
\end{table}

\section{Conclusions}\label{sec:conclusion}
In this work, we propose a simple but effective SVM2+ objective function by utilizing the squared hinge loss instead of the hinge loss
as in the traditional SVM1+ method, which leads to up to hundred times speedup for training the SVM+ classifiers on the
image categorization datasets. In the future, we would like to study the convergence rate and theoretical bound for our proposed SVM2+.

\bibliographystyle{named}
\bibliography{mybib_full}

\end{document}